\documentclass[pdflatex, sn-nature]{sn-jnl}

\usepackage{graphicx}%
\usepackage{multirow}%
\usepackage{amsmath,amssymb,amsfonts}%
\usepackage{amsthm}%
\usepackage{mathrsfs}%
\usepackage[title]{appendix}%
\usepackage{xcolor}%
\usepackage{textcomp}%
\usepackage{manyfoot}%
\usepackage{booktabs}%
\usepackage{algorithm}%
\usepackage{algorithmicx}%
\usepackage{algpseudocode}%
\usepackage{listings}%
\usepackage{placeins}
\usepackage{multirow}
\usepackage{lineno}

\theoremstyle{thmstyleone}%
\theoremstyle{thmstyletwo}%

\theoremstyle{thmstylethree}%

\begin{document}

\title[Article Title]{Single Neuromorphic Memristor closely Emulates Multiple Synaptic Mechanisms for Energy Efficient Neural Networks}


\author*[1]{\fnm{Christoph} \sur{Weilenmann}}\email{weilenmc@iis.ee.ethz.ch}

\author[1]{\fnm{Alexandros} \sur{Ziogas}}

\author[1]{\fnm{Till} \sur{Zellweger}}

\author[1]{\fnm{Kevin} \sur{Portner}}

\author[1]{\fnm{Marko} \sur{Mladenovic}}

\author[1]{\fnm{Manasa} \sur{Kaniselvan}}

\author[2]{\fnm{Timoleon} \sur{Moraitis}}

\author[1]{\fnm{Mathieu} \sur{Luisier}}

\author[1]{\fnm{Alexandros} \sur{Emboras}}

\affil[1]
{Integrated Systems Laboratory\unskip, 
    ETH Zurich\unskip, 8092 Zurich\unskip, Switzerland}
\affil[2]{Huawei Zurich Research Center \unskip, Zurich\unskip, Switzerland}


\abstract {Biological neural networks do not only include long-term memory and weight multiplication capabilities, as commonly assumed in artificial neural networks, but also more complex functions such as short-term memory, short-term plasticity, and meta-plasticity – all collocated within each synapse. Here, we demonstrate memristive nano-devices based on SrTiO\textsubscript{3} that inherently emulate all these synaptic functions. These memristors operate in a non-filamentary, low conductance regime, which enables stable and energy efficient operation. They can act as multi-functional hardware synapses in a class of bio-inspired deep neural networks (DNN) that make use of both long- and short-term synaptic dynamics and are capable of meta-learning or "learning-to-learn". The resulting bio-inspired DNN is then trained to play the video game Atari Pong, a complex reinforcement learning task in a dynamic environment. Our analysis shows that the energy consumption of the DNN with multi-functional memristive synapses decreases by about two orders of magnitude as compared to a pure GPU implementation. Based on this finding, we infer that memristive devices with a better emulation of the synaptic functionalities do not only broaden the applicability of neuromorphic computing, but could also improve the performance and energy costs of certain artificial intelligence applications.

}




\maketitle

\section{Introduction}\label{sec1}

\begin{figure}[hbt!]%
\centering
\includegraphics[width=0.9\textwidth]{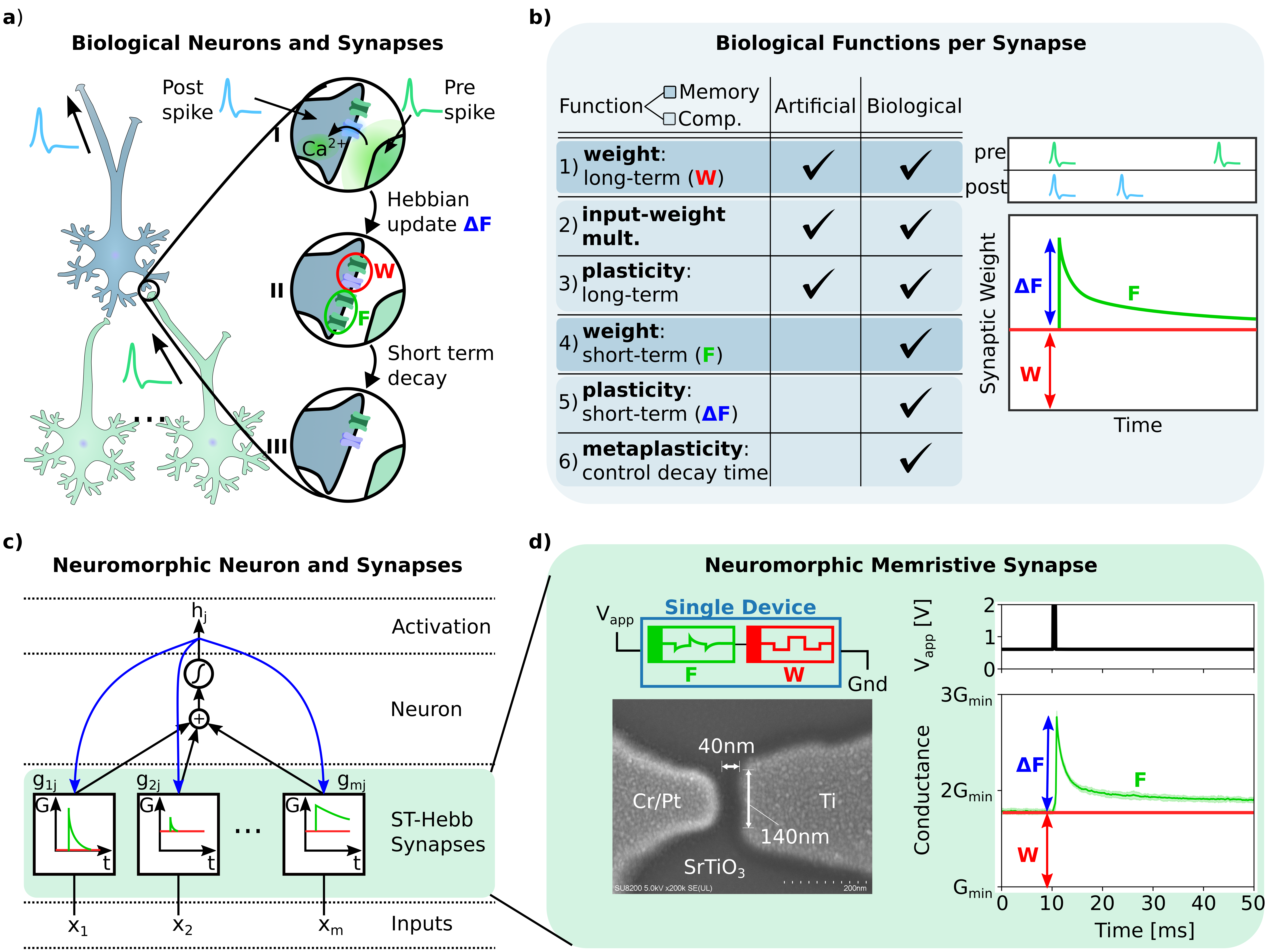}
\caption{Biologically inspired synaptic functions and their memristor implementation. \textbf{a)} Organisation of the mammalian brain with several biological neurons connected through synapses. When a postsynaptic spike (light blue) coincides with a presynaptic spike (light green) the corresponding synaptic coupling is strengthened (Hebbian plasticity) for a limited amount of time (short-term plasticity). This bio-physical process is illustrated in the circular insets: (I) an influx of ions (e.g. $Ca^{2+}$) through the postsynaptic voltage gated ion channels leads to (II) an increased number of synaptic receptors, which increases the synaptic weight. (III) The weight subsequently decays back to its original value due to the receptors gradually detaching from the membrane. \textbf{b)} Table comparing the synaptic functions of artificial synapses in standard ANNs ("Artificial" column) and biological synapses ("Biological" column). The plot on the right shows the weight of a biological synapse as a function of time. The short-term weight (F) is updated ($\Delta F$) when the pre- and post-synaptic spikes coincide. Additionally, the decay time of F can be controlled, which corresponds to meta-plasticity. \textbf{c)} Bio-inspired Short-Term Plasticity Neuron (STPN) model combining a conventional neuron model with short-term Hebbian (ST-Hebb) synapses. \textbf{d)} Hardware implementation of a neuromorphic ST-Hebb synapse with a Cr/Pt-SrTiO\textsubscript{3}-Ti memristor. The device measurement on the right mirrors the biological functions of ST-Hebb combining memory and computation as well as long- (W) and short-term (F) dynamics.}\label{fig1}
\end{figure}

Biological neural networks (BNNs) have been used as source of inspiration for today's most successful artificial neural networks (ANNs). Specifically, ANNs abstract the brain's complex functionality into a network graph where the nodes represent neurons and the connections synapses. Each synapse in such a network traditionally has three functions: (1) storage of long-term memories in their weights (W), (2) synaptic transmission – modelled as input-weight multiplication, and (3) long-term plasticity - the update of W during training. However, such a model only captures a sub-set of the functionalities of biological synapses, which obey complex dynamics and learning rules such as Hebbian plasticity \cite{Markram2012Spike-timing-dependentOverview} and short-term plasticity \cite{Erickson2010AMemory, short-term_zucker} (Fig. \ref{fig1}a). In addition, higher-order plasticity rules exist. They do not directly determine the synaptic weight, but rather the properties of the plasticity rule itself. One example is the control over the decay timescale of the short-term plasticity rule, which can range from milliseconds to minutes, depending on the neuronal activation \cite{wang2006heterogeneity, erickson2010single}. Such rules have been termed meta-plasticity \cite{abraham1996metaplasticity, Barrett2009StateCapture}. They play an important role in demanding tasks where not only learning is necessary, but also learning-to-learn, i.e., meta-learning \cite{finn2017model, miconi2018differentiable, miconi2020backpropamine, Tyulmankov2022Meta-learningDetection, Najarro2020Meta-learningNetworks, Hospedales2022}. This complexity of synaptic biophysics and the corresponding plasticity rules are crucial for the function of the nervous system (e.g., \cite{nadim2000role, citri2008synaptic, shimizu2021computational}), but they are lacking in conventional ANNs. The resulting limited biological realism could partly explain the inferior performance of artificial intelligence (AI) systems compared to humans and animals in many aspects, such as motor skills and adaptability to dynamic environments \cite{zador2023catalyzing}. Moreover, today's ANNs consume vast amounts of energy due to the large network size required for complex tasks \cite{Canziani2016AnApplications}. The training of the large language model behind ChatGPT (GPT3) consumed for example, 1.287 GWh of electrical energy \cite{Patterson2021CarbonTraining}, enough to power more than 100 households for a year. \\

To address these issues associated with current ANNs, a more bio-inspired synaptic model was conceived that includes short-term and Hebbian plasticity as well as meta-plasticity. This model, referred to as ST-Hebb synapse, was shown to optimally adapt to continuous, dynamic environments  \cite{Moraitis2020Short-termEnvironments}. Specifically, the ST-Hebb synapse does not only perform the aforementioned three functions of traditional ANN synapses, but also includes the following additional roles (Fig. \ref{fig1}b): (4) storage of short-term memories (F) that decay over time, (5) short-term plasticity - the update of F ($\Delta F$) during training and inference, and (6) meta-plasticity - the control over the decay time. To make use of the ST-Hebb synapse in a deep neural network (DNN), the short-term plasticity neuron (STPN) model shown in Fig. \ref{fig1}c has been recently proposed \cite{Rodriguez2022Short-TermForget}. It combines a conventional neuron model with ST-Hebb synapses. Importantly, this model makes use of all six synaptic functions (1) to (6) incorporates meta-learning, it can be integrated into multi-layer networks, and it outperforms more conventional ANNs with less biologically realistic synapses in a variety of challenging tasks. 


The hardware of choice to run such neural networks are parallel computing architectures like graphics processing units (GPUs). However, GPU-based implementations of multi-functional synapses suffer from the computational overhead caused by the aforementioned additional synaptic operations. This trend is exacerbated by the large amount of synapses building state-of-the-art neural networks, ranging from $10^{6}$ to $10^{14}$ \cite{Xu2018ScalingNetworks}. On top of that, the operations governing ST-Hebb’s synaptic dynamics are memory bound and are thus negatively affected by the well-known von Neumann bottleneck imposed by physically separated memory and processing units \cite{Yu2018Neuro-InspiredMemorys}. These factors render the implementation of ST-Hebb synapses on GPUs inefficient, thus motivating the development of new hardware paradigms that are better suited to neural networks with multi-functional synapses. \\

Several promising neuromorphic architectures use memristors as hardware synapses because of their ability to collocate memory and computation in a single device, which circumvents the von Neumann bottleneck \cite{Sebastian2020}. Memristors are two-terminal devices that can change their conductance state upon electrical \cite{Jo2010, waser} or optical \cite{Emboras2020, Portner2021} stimuli, similar to the change of the synaptic coupling (weight) upon a neuronal spike in biological systems. A growing body of research suggests that the rich internal dynamics of memristors can be leveraged to mimic biophysical processes taking place in synapses and neurons  \cite{Kumar2022DynamicalComputing, Demirag2021PCM-trace:Materials}. 

There have been multiple demonstrations of bio-inspired hardware synapses realized using memristors with both long- and short-term dynamics \cite{Yang2019MemristiveComputing, Choi2020EmergingComputing} that exhibit biological learning rules such as triplet spike-timing-dependent plasticity (triplet-STDP) \cite{Yang2018} or Bienenstock-Cooper-Monroe (BCM) \cite{Xiong2019}. However, these demonstrations rely on spike timing plasticity rules and can therefore not be integrated into DNNs \cite{Pfeiffer2018}, which limits their applicability. Meanwhile, a single-layer neural network that makes use of bio-inspired, multi-functional synapses was recently demonstrated on memristive hardware \cite{Sarwat2022Phase-changeComputations}. The authors showed the benefit of adding short-term synaptic plasticity during inference for a classification task in dynamically changing environments. Memtransistive devices were used as synapses. In addition to the two electrical contacts common to all memristors, they possess a gate analogous to transistors. To realize decaying traces a voltage signal with the shape of the short-term decay was applied to the third gate contact. Short-term plasticity is therefore not an intrinsic property of these devices, i.e., the devices do not inherently exhibit short-term memory, but require an additional stimulus to do so. The need for three-terminal devices and precisely engineered voltage signals applied to each memtransistive synapse poses challenges for a large-scale implementation of such systems, because the required control circuit and wiring would rapidly become considerably complex. Therefore, the introduction of a two-terminal memristive device that intrinsically encompasses all six synaptic roles (1-6) is key to enable scalable neuromorphic hardware that is not only energy-efficient, but also reaches or even surpasses the performance of conventional AI approaches.\\


In this work, we propose such a two-terminal memristive device that relies on the valence-change switching mechanism in SrTiO\textsubscript{3} (STO) and intrinsically possesses the six operations needed to function as an ST-Hebb synapse. A symbolic representation on top of an SEM image of the fabricated nanoscale device is shown in Fig. \ref{fig1}d. The measured memristor conductance acts as the plastic synaptic weight and mirrors the behavior displayed in Fig. \ref{fig1}b. Specifically, our device can store two different states in its memory, (I) a state with slow dynamics (long-term weight W) and (II) a state with fast dynamics (short-term weight F), which are both encoded in the conductance of the memristor. In terms of computation, the four synaptic operations labelled 2, 3, 5 and 6 in Fig. \ref{fig1}b can all be performed by our STO devices: (III) Long-term plasticity (i.e., change in the long-term weight W) and (IV) short-term plasticity (short-term weight update $\Delta F$) can both be triggered by voltage pulses of different magnitudes. Notably, the short-term decay happens spontaneously, without the application of a complex signal. (V) Meta-plasticity (i.e., control over the decay time) can be achieved by applying a DC bias voltage to one of the two terminals, which limits the complexity of the control circuit and wiring. (VI) Additionally, our devices provide the standard in-memory multiplication capabilities of the input (voltage U) by the synaptic weight (conductance G), which is realized by Ohm’s law $I = G \cdot U$. They also exhibit low cycle-to-cycle variability due to their non-filamentary switching operation. As a consequence, the random displacement of few atoms does not induce as much noise as in filamentary valence-change-type memristors \cite{stochastic_filament}. Moreover, we can operate our devices at very low conductance values (10s of nS), which lowers the power consumption during operation. Their achievable short-term timescales range from 10 milliseconds to 100’s of seconds. Importantly, timescales in the order of 100 seconds are typically difficult to realize with nanoscale footprints using other neuromorphic approaches such as analog circuits because the required capacitors rely on much larger dimensions  \cite{Cruz-Albrecht2012Energy-efficientCircuits, Joubert2012HardwareDigital, Gopalakrishnan2015TripletImplementation}.
\\

To estimate the energy consumption of our multi-functional hardware synapses in the context of a large DNN, we introduce a modified STPN (m-STPN) unit that emulates parts of the device characteristics and fully incorporates the measured energy consumption of our devices. We then integrate this unit into the original STPN network simulator of \cite{Rodriguez2022Short-TermForget} to perform a complex reinforcement learning task in software with multi-functional synapses, namely learning to play Atari’s video game Pong. The Atari suite is a common benchmark for reinforcement learning and is chosen here as an exemplary task for a dynamic environment. We show that the m-STPN unit enables faster and more stable training compared to the original version. The main reason for this is the introduced constraint on the short-term decay time constant imposed by our devices. Furthermore, we demonstrate that short-term weights with long timescales, such as the ones exhibited by our memristors, are required for a robust and fast training of the network. Finally, we compare the network's energy consumption for a pure GPU implementation of the synapses with the estimated energy consumed by our memristive synapses. We demonstrate an estimated gain in energy efficiency between 96$\times$ and 970$\times$, depending on the GPU implementation.

\FloatBarrier

\section{Multi-functional Synaptic Behavior in Single Memristor}\label{sec2}

We fabricated a multi-functional memristive synapse on an STO single crystal substrate (Fig \ref{fig1}d). First, a high work function contact (Pt with a Cr adhesion layer beneath) was deposited. This step was followed by the fabrication of a Ti electrode with a Pt capping layer that prevents the Ti from oxidizing in air. Both contacts were deposited using electron beam evaporation and patterned by electron beam lithography with a subsequent lift-off process, resulting in a typical gap between the electrodes of roughly 40nm. The devices were annealed at 300°C for 20min in flowing Ar, which causes a thermal oxide to form at the Ti-STO interface. The whole stack was finally covered with a uniform layer of 15nm of SiN. The fabrication process is discussed in detail in Methods Section \ref{sec:device_fab}.

\begin{figure}[h!]%
\centering
\includegraphics[width=0.9\textwidth]{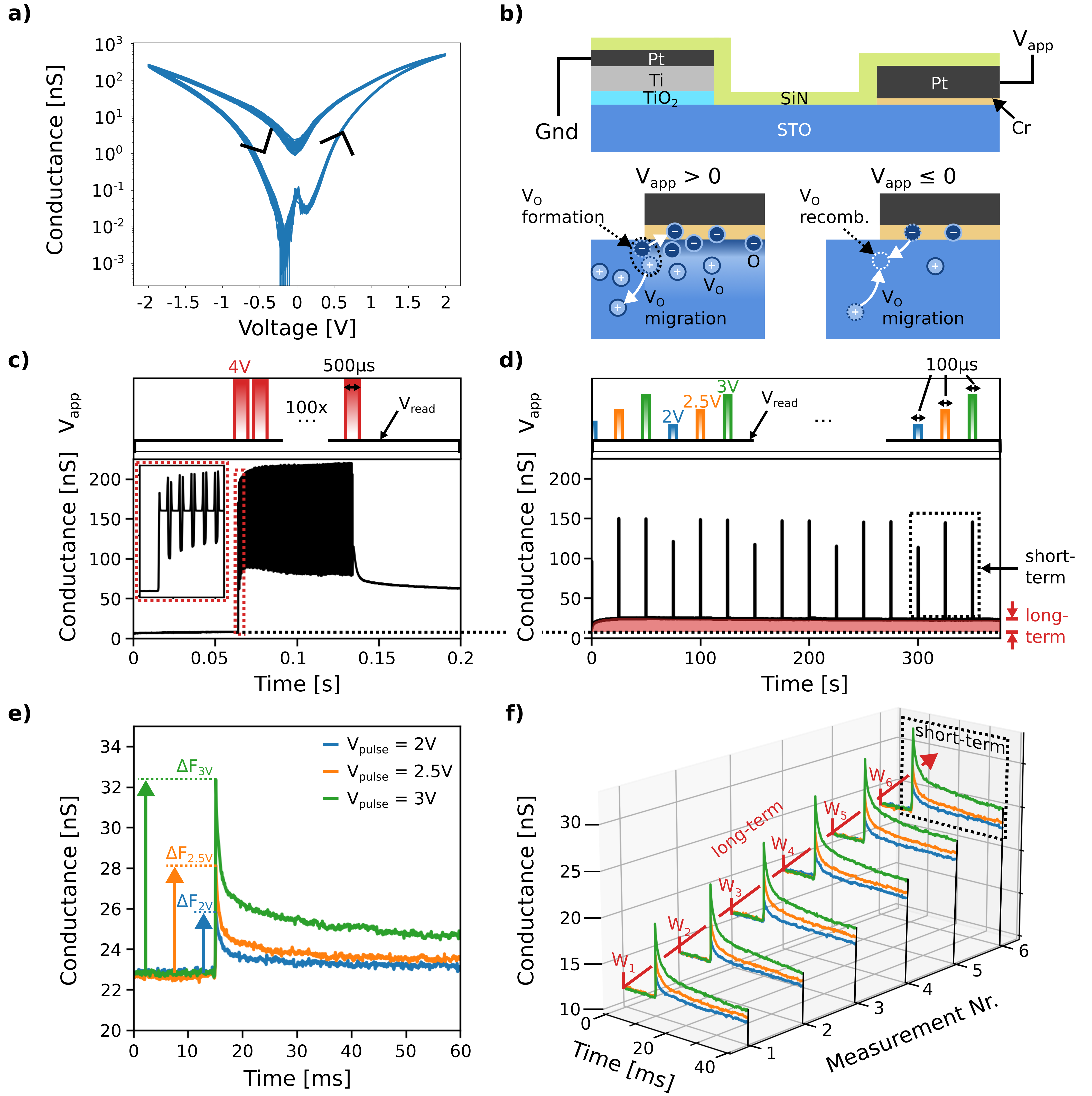}
\caption{DC and dynamical behavior of multi-functional memristive synapses. \textbf{a)} Conductance vs. voltage characteristic of the fabricated Cr/Pt-STO-Ti memristors. The black arrows indicate the counter-clockwise switching direction. \textbf{b)} Sketch of the device stack and of the underlying switching mechanism. The two insets zoom into the Pt-STO interface at different applied voltages, showing the dynamics of interfacial oxygen ions (O) and oxygen vacancies ($V_O$): ($V_{app} > 0$) At positive voltages $V_O$ formation and migration occurs. The negatively charged oxygen migrates towards the interface and into the porous Pt electrode and the positively charged $V_O$ move along the electric field away from the Pt electrode and towards the grounded Ti electrode. ($V_{app} \leq 0$) At zero applied voltage,  $V_O$'s move back towards the Pt, driven by the built-in electrochemical gradient, where they recombine with O. A negative voltage accelerates this process. \textbf{c)} Conductance change from low to high under the application of 100 SET pulses with an amplitude of 4V and a duration of 500 $\mu$s. \textbf{d)} Time-dependent conductance measurement (read out at 0.6V) when a series of voltage pulses with an amplitude of 2V, 2.5V, and 3V and a duration of 100$\mu$s are applied. The long-term conductance (red area) remains constant. The pulses induce short-term increases of the conductance with subsequent decay. \textbf{e)} Short-term conductance changes due to the last three voltage pulses of the measurement in d), indicated by the dotted rectangle. Only the conductance values during the read voltage are shown here. \textbf{f)} Aggregate plot showing short-term plasticity for different values of the long-term weight W. The measurement data was obtained by first applying the protocol in d) to characterize the short-term plasticity for the minimum long-term weight (W\textsubscript{1}). The long-term weight was then changed by 100 SET pulses (c) and, after a waiting period of 240s, the short-term plasticity was measured again.}\label{fig2}
\end{figure}

Figure \ref{fig2}a shows 30 cycles of the I-V characteristics of our Cr/Pt-STO-Ti memristor (Fig. \ref{fig2}b). The voltage (-2V to 2V) is applied to the Pt electrode, while the Ti one is grounded. A high cycle-to-cycle repeatability as well as low conductance values (100s of nS) are obtained, which allows for energy-efficient device operation. The low conductance values and the counter clockwise switching direction, as indicated by the black arrows, are attributed to a non-filamentary switching mechanism, which has already been reported for similar material stacks \cite{Muenstermann2010CoexistenceDevices}. In this switching regime the conductance change is not caused by the formation of a filament made of oxygen vacancies ($V_O^{2+}$) that bridges the two electrodes, but by the modulation of the Schottky barrier at the Pt-STO interface \cite{Baeumer2016}. This modulation is attributed to generation and recombination of $V_O^{2+}$'s upon the application of an external voltage (bottom of Fig. \ref{fig2}b). The vacancies in turn locally dope the STO, which changes the height and width of the Schottky barrier, affecting the conductance. When a positive voltage is applied to the Pt contact, oxygen ions ($O^{2-}$) from the crystal move to the Pt-STO interface or into the porous Pt electrode, leaving behind a positively charged crystal defect \cite{Cooper2017AnomalousTEM}. This kind of n-type doping increases the conductance due to a decrease in the Schottky barrier height and width. Since $V_O^{2+}$'s are mobile and positively charged, they migrate away from the Pt electrode along the applied electric field towards the Ti electrode, where they accumulate and potentially form a filament in a process called electroforming \cite{Menzel2019MechanismOxRAM}. We observed that for high positive voltages ($> 4V$) we are able to electroform our device and put it in a filamentary-switching operation (Supplementary Section S2). This confirms the generation of $V_O^{2+}$'s at positive voltages in our devices. The amount of vacancies generated as well as the distance over which the $V_O^{2+}$'s migrate from the Pt contact depend on the voltage and duration of the applied electrical signal \cite{Siegel2021}. When no external voltage is applied, the vacancies diffuse back towards the Pt contact, driven by a gradient in electrochemical potential and recombine there with the interfacial oxygen. This gradually resets the Schottky barrier and thus the conductance. This process is aided by the application of a negative bias, leading to a faster conductance decay. Hence, the decay time can be voltage-controlled. Even though this physical picture explains our experimental observations it cannot be excluded that trap states at the interface play a key role in the switching process, as suggested by other works \cite{Mikheev2014ResistiveJunctions}. \\

In our approach the memristor's conductance implements the synaptic weight, whose dynamics (long- and short-term) are crucial in ST-Hebb synapses (Fig. \ref{fig1}b). To investigate the conductance dynamics of our STO memristors we apply pulses of different voltages and widths  to them (Figs. \ref{fig2}c and \ref{fig2}d). We first induce long-term plasticity (function 3 in Fig. \ref{fig1}b) by applying 100 SET pulses with an amplitude of 4V and a duration of 500$\mu$s that cause the device to switch from a low to a high conductance state (Fig. \ref{fig2}c). This high conductance state slowly decays over thousands of seconds without applied bias (Supplementary Section S3). After the SET procedure we leave the device at 0V for 240s (not shown) to let it settle to a stable state. We then proceed with measuring the conductance of the device at 0.6V for 375s (Fig. \ref{fig2}d), during which 100$\mu$s-long pulses with voltages of 2, 2.5, and 3V, are applied. The long-term conductance induced by the SET pulses remains largely constant for the time period of the measurement. The 100$\mu$s-long pulses lead to a short term conductance increase, i.e. short-term plasticity (function 5 in Fig. \ref{fig1}b), whose magnitude depends on the pulse voltage (3, 5 and 10nS for 2, 2.5 and 3V, respectively) and is followed by a decay. This can be observed in Fig. \ref{fig2}e, where the conductance during the three last pulses of the protocol (dotted rectangle in Fig. \ref{fig2}d)  is plotted. The conductance during the read voltage is shown, omitting the values during the 100$\mu$s-long pulse.
In Fig. \ref{fig2}f the long- and short-term components of the conductance (functions 1 and 4 in Fig. \ref{fig1}b) are visualized for six measurements with different values of the long-term weight W. The measurement data was obtained by repeating the protocol of Fig. \ref{fig2}c and \ref{fig2}d multiple times, i.e., first setting the long-term weight (W\textsubscript{1}, W\textsubscript{2}, ...) by 100 SET pulses, waiting for 240s, and then applying the short-term pulse protocol of Fig. \ref{fig2}d. Values of the long-term conductance in the range of 12 to 23nS can be set in this way (long-term plasticity). These conductance values can further undergo short-term increases induced by voltage pulses (short-term plasticity). The obtained collocation of both long- and short-term plasticity motivates the use of this devices as ST-Hebb synapses.\\

\FloatBarrier

\begin{figure}[h!]%
\centering
\includegraphics[width=0.9\textwidth]{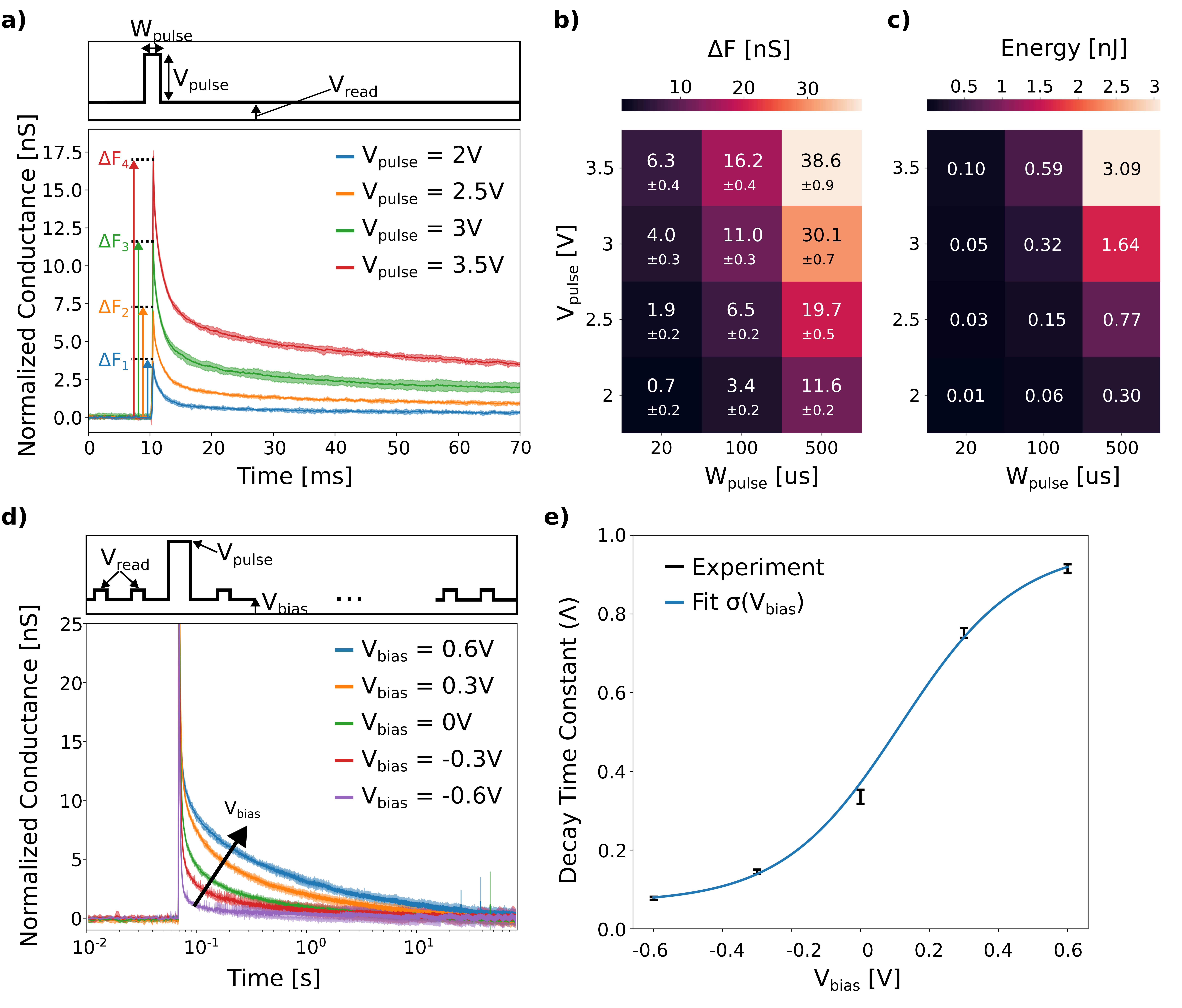}
\caption{Control over the magnitude and dynamics of short-term conductance updates. \textbf{a)} Mean (solid line) and standard deviation (shaded area) of pulse-induced short-term conductance updates ($\Delta F$) from five conductance measurements and using four different pulse voltages (2, 2.5, 3, and 3.5V). The read voltage is set to 0.6V and the pulse width to 100 $\mu$s. To better compare the measurements, the conductance values were adjusted by subtracting the initial conductance at t=0 from the data. \textbf{b)} Heatmap of the achieved $\Delta F$ for the different pulse voltages and widths. \textbf{c)} Heatmap of the required pulse energy for the same voltage and width combinations as in b). \textbf{d)} Applied voltage protocol on a linear x-axis (top) and corresponding conductance values using 0.6V read pulses shown on a logarithmic x-axis (bottom). In between the read pulses a constant bias voltage ($V_{bias}$) of variable amplitude is applied. The main pulse voltage and width are set to 3.5V and 500 $\mu s$, respectively, in all measurements. The mean and standard deviation of the adjusted conductance values are shown for 5 measurements on a semi-log plot. \textbf{e)} Extracted decay time constant $\Lambda$ from the measurements in d) as a function of $V_{bias}$. The experimental data points were fitted with a sigmoid function.}\label{fig3}
\end{figure}

The short-term plasticity is investigated in more detail in Figure \ref{fig3}a, which displays the mean (solid line) and standard deviation (shaded area) of five measurements. Pulse-induced short-term conductance updates ($\Delta F$) and subsequent decays are obtained using four different voltage amplitudes (2, 2.5, 3, and 3.5V). The pulse width was fixed to 100$\mu$s and the read voltage to 0.6V. The conductance values were normalized by subtracting the initial conductance at t=0 from the data. We observe low cycle-to-cycle variability, in agreement with the I-V characteristics in Fig. \ref{fig2}a. The same measurement was repeated for two additional pulse widths (20 and 500 $\mu$s). The resulting $\Delta F$'s are reported in Fig. \ref{fig3}b as a function of the pulse amplitude and width. It can be seen that $\Delta F$ values in the range of 0.7 - 38.6 nS can be achieved by adjusting these parameters. The corresponding energy per pulse is given in Fig. \ref{fig3}c for the same pulse voltage and width combinations. The details of the energy calculations are given in Supplementary Section S4.
\\
Besides the magnitude of the conductance increase, it is also possible to control the subsequent decay using a DC bias voltage ($V_{bias}$) that is constantly applied during the experiment (Fig. \ref{fig3}d), effectively implementing meta-plasticity (function 6 in Fig. \ref{fig1}b). The mean and standard deviation of the conductance for five measurements are shown as a function of time. The voltage pulse that triggers the conductance increase is the same in all cases, thus resulting in similar $\Delta F$, whereas the bias voltage is varied (see Supplementary Section S6 for details). The timescale of the decay increases with increasing $V_{bias}$ from hundreds of ms ($V_{bias} = -0.6V$) to tens of seconds ($V_{bias} = 0.6V$). To quantify the resulting decay time constant ($\Lambda$) as a function of the bias voltage, we fitted an exponential to the measured curves (Supplementary Section S6). In our fit, the maximum value of $\Lambda = 1$ indicates no decay and the minimum value ($\Lambda = 0$) corresponds to immediate decay. Figure \ref{fig3}e demonstrates that we can experimentally control $\Lambda$ over a range from 0.08 to 0.92 as a function of the applied $V_{bias}$. The relationship between $V_{bias}$ and $\Lambda$ is modelled by a sigmoid function $\Lambda (V_{bias}) = \frac{L}{1+\exp(-k\cdot(V_{bias}-V_0))} + \Lambda_0$, where L, k, $V_0$, and $\Lambda_0$ are fitting parameters.
\\
In summary, the following functions are performed intrinsically by our memristors: Storing both (1) long- ($\textbf{W}$) and (2) short-term ($\textbf{F}^{(t)}$) weights (Fig. \ref{fig2}f), (3) long-term plasticity (Fig. \ref{fig2}c), (4) short-term plasticity (Figs. \ref{fig3}a and \ref{fig3}b), (5) meta-plasticity via control over the decay time parameter $\boldsymbol{\Lambda}$ (Figs. \ref{fig3}d and \ref{fig3}e), and (6) multiplication of the input voltage with the synaptic weight according to Ohm's law.

\FloatBarrier

\section{DNN with Multi-functional Memristive Synapses}

The six intrinsic functionalities of our memristors can be utilized by ST-Hebb synapses in a deep STPN network.  Such networks have been shown to outperform traditional DNN implementations without multi-functional synapses at a variety of complex tasks in dynamic environments \cite{Rodriguez2022Short-TermForget} . One such dynamic task is learning to play Atari Pong, a video game and common machine learning benchmark. In Pong a player (the STPN network) confronts an opponent, each manipulating a vertically movable bar to strike a ball, aiming to get the ball past the opponents bar (i.e., scoring a point) or preventing the opponent from doing so. The game concludes when either player scored 21 points. The STPN network's reward is the difference between the player's and the opponent's points at the end of the game. Given only this scalar reward as input, the network finds a strategy that results in the maximum score of 21 by repeatedly playing the game and employing reinforcement learning, a bio-inspired learning paradigm \cite{Neftci2019ReinforcementSystems}.
Below we show the development of a modified STPN unit (m-STPN), an altered version of the one in \cite{Rodriguez2022Short-TermForget}, which makes use of our multi-functional synapses to play Atari Pong. Through simulation we could estimate the energy consumption of the whole network if it were running on our memristive hardware and compare it to a pure GPU implementation.

\subsection{Modified short-term plasticity neuron}

\begin{figure}[h!]%
\centering
\includegraphics[width=0.9\textwidth]{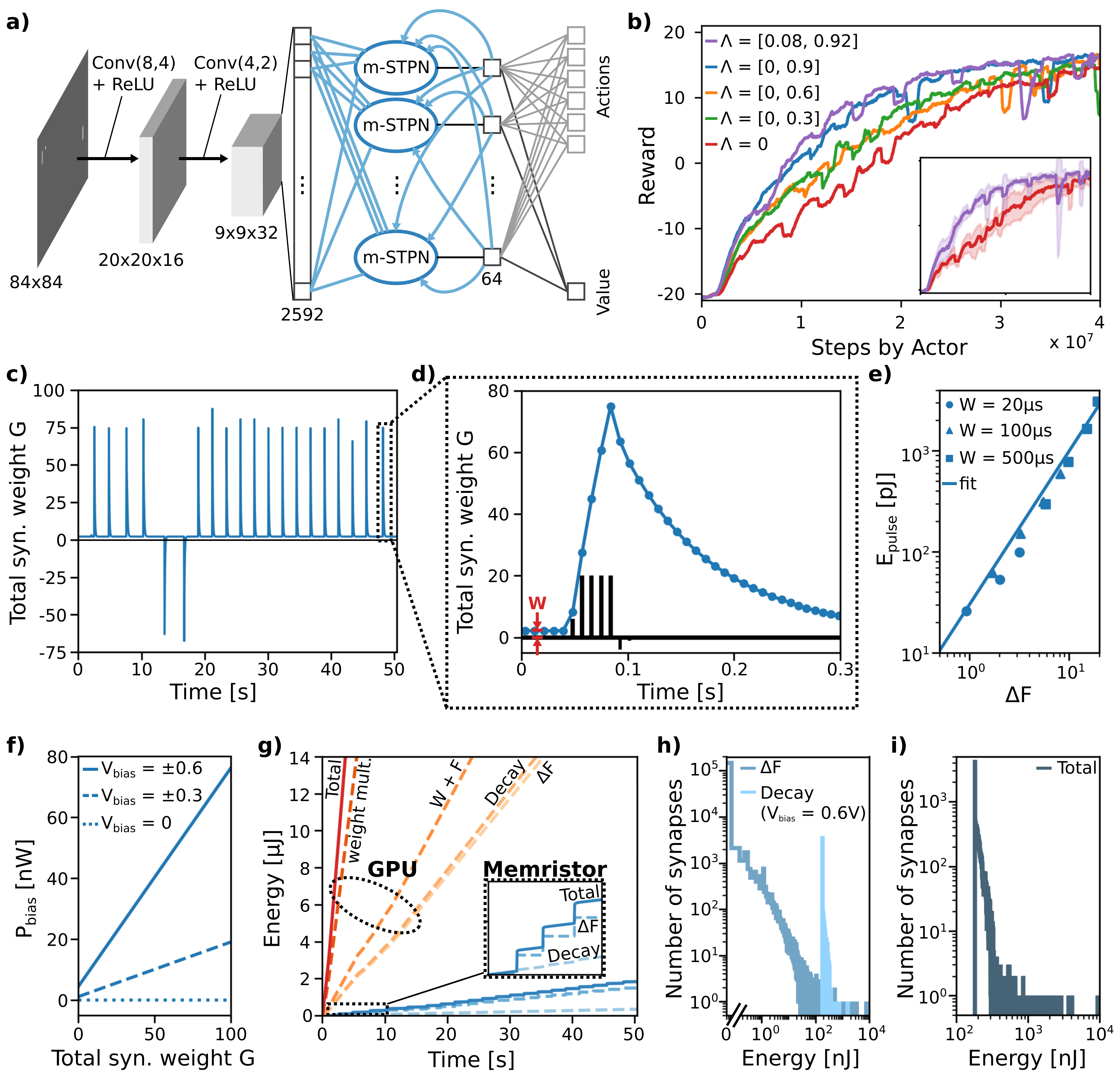}
\caption{Simulation and energy consumption of an STPN network with multi-functional synapses. \textbf{a)} Sketch of the full STPN network. A frame of the Atari game is fed into two convolutional layers: Conv(kernel, stride) plus a ReLU activation function. The features are then flattened into a 1D array and fed into the m-STPN layer consisting of m-STPN units (blue ellipses) and the corresponding ST-Hebb synapses (blue lines). The layer's output is split into "actions" and a "value" by two fully connected linear layers. \textbf{b)} Average reward as a function of the agent steps during training for five different ranges of $\Lambda$. Each curve represents the average reward of 16 agents with different random parameter initialization. In the inset, the cases $\Lambda = 0$ and $\Lambda = [0.08, 0.92]$ (i.e., the achievable device range) are shown with their standard deviation (shaded area). \textbf{c)} Total synaptic weight (long- and short-term component) of a single synapse of the trained network ($S_{max\{\Delta F\}}$) during an entire game. \textbf{d)} Zoom-in of the dotted area in c). The long-term weight W is marked in red. The black bars represent the $\Delta F$ at each time step. \textbf{e)} Experimental values of the energy consumed by the voltage pulses required to change the short-term weight as a function of $\Delta F$. These values follow a power-law relationship. \textbf{f)} Calculated power consumption of a single synapse as a function of the total synaptic weight at different $V_{bias}$ used to vary the short-term decay constant. \textbf{g)} Time evolution of the estimated energy consumption of the synapse in c) during an entire Pong game (roughly 50 seconds) comparing a memristive synapse with a pure GPU implementation. The different components as well as the total energy consumption are shown. \textbf{h)} Histograms of all synapses in the network, indicating how many synapses consume a specific amount of energy during the whole Pong game. The two different components of the energy consumption ($\Delta F$ and Decay) are shown analogous to g). A worst case scenario of $V_{bias} = 0.6$ for all synapses is assumed. \textbf{i)} Total energy ($\Delta F$ plus Decay) consumed per synapse for the entire game.}\label{fig4}
\end{figure}

The deep STPN network simulator investigated here (Fig. \ref{fig4}a) employs our modified STPN units (m-STPN) as described in detail in Methods Section \ref{sec:mSTPN}. The network relies on an actor-critic architecture that takes frames of the Atari Pong environment as inputs and computes both the next action to take in the environment (actor) as well as an estimation of the value of the current state (critic). The frames are first processed by two convolutional layers into a dense feature set that forms the input for the m-STPN layer. The latter consists of 64 m-STPN units, each of which is connected through ST-Hebb synapses to the 2592 inputs as well as recurrently to 64 outputs. In total, this amounts to $(2592+64)\cdot 64 = 169984$ synapses. The output of the m-STPN layer is then fed into two fully-connected linear layers that compute the next action (the actor’s next step to take in the game) and the current value (how advantageous is the current game state).
Our aim is to show that our mutli-functional memristors can act as hardware ST-Hebb synapses in the STPN network of Fig. \ref{fig4}a. To achieve this, the following device characteristics where implemented into the m-STPN units: (1) mapping of the memristor conductance ($G_{meas}$) to the simulated, unitless synaptic weight ($G$) by the linear relationship 
\begin{equation}
    G = (G_{meas}-G_{min}) / m
    \label{eq4}
\end{equation}
with $m=2nS$ and $G_{min} = 12nS$. (2) Adding a discretization operation to the simulated short-term weight update ($\Delta F$) that limits the number of $\Delta F$ values (states) to an amount that can be resolved by our memristors. To satisfy this requirement, the conductance values corresponding to two adjacent states should be separated by at least one standard deviation, which is below 1nS for all short-term weight updates $\Delta F_{meas}$ (max. $\pm 0.9$ nS in Fig. \ref{fig3}b). We therefore chose a discretization step of 1nS for $\Delta F_{meas}$, which translates to a step of 0.5 for the simulated $\Delta F$ according to Eq. \ref{eq4}. (3) Fixing the maximum of $|\Delta F|$ to 20, which makes sure that the weight update remains in a range that is achievable by the STO memristors. A histogram of $\Delta F$ for all synapses during an entire Pong game, with and without non-idealities, is given in Supplementary Section S8. (4) Limiting the range of the decay time constant $\Lambda$ to values that can be reached by our devices ([0.08, 0.92]). The constraining of $\Lambda$ has an impact on the training performance of the network, as shown in Fig. \ref{fig4}b. The reward during training is plotted as a function of the steps taken by the actor (see Methods Section \ref{sec:training} for details). Each curve represents the average reward of 16 agents that learn to play the game with different randomly initialized parameters. The five lines denote different constraints imposed on the learned decay time parameter $\Lambda$. Notably, it is beneficial to incorporate synapses with large decay time constants during training: The larger the upper limit of $\Lambda$ the faster the reward increases. Unexpectedly, the case with $\Lambda = 0$ (i.e., immediate decay of the short-term weight changes for all synapses) also learns, albeit slower and less robustly, as can be seen from the larger standard deviation compared to $\Lambda = [0.08, 0.92]$ (inset of Fig. \ref{fig4}b). The longer, constrained decay times were made possible by the modified weight normalization scheme in m-STPN's (Methods Section \ref{sec:mSTPN}). Because the decay constant $\Lambda$ is naturally limited in our devices, destabilizing phenomena such as an exponential gain ($\Lambda > 1$) instead of a decay ($\Lambda < 1$) are automatically prevented.\\ 

After training some of the 16 trained agents achieve the maximum reward of 21 (Supplementary Section S9). The total synaptic weight value $G = W + F$ of a single synapse of such a trained agent is reported in Fig. \ref{fig4}c over the course of an entire game that lasts roughly 50 seconds. This specific synapse was chosen because it exhibits the largest synaptic changes ($\Delta F$) in the whole network. It is therefore referred to as $S_{max\{\Delta F\}}$ in the remainder of the text and will serve as a representative example for the behavior of a synapse in an STPN network. It is observed that the value of the synapse's weight $G$ changes over time due to the short-term plasticity of ST-Hebb synapses. Importantly, the short-term updates are sparse, which makes the implementation of this reinforcement learning task energy efficient on our memristive hardware as only a small number of energy consuming short-term weight updates ($\Delta F$) are needed. Figure \ref{fig4}d presents a zoom-in of the simulated synaptic weight G where both the long-term weight component W (in red) and the short-term weight updates $\Delta F$ (in black) are shown. Each simulation timestep is marked by a dot. 
\FloatBarrier

\subsection{Energy consumption of deep STPN network}

Next, we estimate the energy consumption of synapse $S_{max\{\Delta F\}}$ for the duration of the entire game if it were implemented on our memristor. Two sources of energy loss are considered: Firstly, each voltage pulse that causes a short-term weight update consumes energy ($E_{pulse}$) (Fig. \ref{fig3}c). Secondly, due to the application of a constant bias to control the decay time a small current continuously flows through the devices, inducing a power loss ($P_{bias}$). We address these two components separately. Figure \ref{fig4}e reports the first one ($E_{pulse}$) as a function of the short-term weight updates $\Delta F$. This quantity is extracted from the measurement data in Figs. \ref{fig3}b and \ref{fig3}c, for different pulse widths. The measured energy data points closely follow a power law relation: $E_{pulse}(\Delta F) = c \cdot (\Delta F)^{\alpha}$ with $c=30$pJ and $\alpha = 1.52$. This power law relation was incorporated into our neural network simulator to estimate the energy consumption of the short-term weight updates in our memristors. Because the value of $|\Delta F|$ is limited to 20 and because the weight updates are sparse, this first contribution to the energy consumption remains low. In Fig. \ref{fig4}f the second contribution to the energy consumption ($P_{bias}$) is given as a function of the total synaptic weight $G$. It is calculated according to $P_{bias} = |G_{meas}| \cdot V_{bias}^2$. Note that even for a simulated weight of $G = 0$ there is a remnant power draw (except if $V_{bias} = 0$) because of the finite minimum conductance value $G_{min} = 12nS$ of the physical devices. For the maximum bias voltage $V_{bias} = 0.6$ the power consumed by a synapse with a constant weight of $G = 0$ is therefore 4.3 nW. This low power consumption is a direct consequence of our memristor's low conductance values, enabled by their non-filamentary switching behavior.

In Fig. \ref{fig4}g the estimated energy consumed during inference over the course of a Pong game by either a memristor (blue) or a pure GPU implementation (orange) of synapse $S_{max\{\Delta F\}}$ is provided. In the memristor case, the energy consumption can be decomposed into two contributions, the short-term weight updates ($\Delta F$) and the applied bias voltage needed to control the decay time constant (Decay). These two components cover the short-term synaptic plasticity and meta-plasticity required by an ST-Hebb synapse during inference. The standard input-weight multiplication is obtained through Ohm's law $I=G\cdot V_{read}$, where $V_{read}$ encodes the input. The power consumed by this operation is however already accounted for by $P_{bias}$: The current resulting from the application of the maximum bias voltage $\max \{V_{bias}\} = 0.6V$ can be read out to compute the input-weight multiplication. To implement the same plasticity, meta-plasticity, and input-weight multiplication on a GPU the following four operations need to be executed at every time step during the game (6826 in total): (1) Element-wise addition of short- and long-term weight components, (2) element-wise multiplication of F with $\Lambda$ for the short-term decay, (3) element-wise addition of F and $\Delta F$ for the short-term weight update, and (4) vector-matrix multiplication of inputs and weights (weight mult.). For each of these operations the GPU's energy consumption was measured for a single synapse (see Methods Section \ref{sec:GPU_energy}). It is found that the energy consumption of the memristor increases more slowly with the number of timesteps than the GPU baseline. It should however be noted that even though our multi-functional memristive synapse can fully mimic the behavior of an ST-Hebb synapse, the operations of the neuron still need to be performed on a GPU: This concerns the calculation of the magnitude of $\Delta F$ via the first term in Methods Eq. \ref{eq3}, the calculation of the non-linear activation function in Methods Eq. \ref{eq2}, and the normalization of the pre-synaptic input (Supplementary Fig. S7b). \\

To estimate the total synaptic energy consumption of the whole network the contribution of each synapse for an entire game of Pong has to be considered (Fig. \ref{fig4}h). Both the energy consumed by the $\Delta F$ updates (dark blue), and by the control of the decay time constant (light blue) are shown in the form of a histogram. Most synapses do not undergo any short-term weight update during the entire game and therefore do not consume energy for this operation, as indicated by the large $\Delta F$ spike centered around 0. For the decay control, we assume the worst-case scenario where a bias voltage of 0.6 V is applied to all synapses. The current due to this bias can be read out, which accounts for the energy consumption due to the calculation of the vector-matrix multiplication between the input and the weights. A crossbar array architecture is assumed for this purpose. 

The total energy (i.e., $\Delta F$ plus Decay) consumed by each memristive synapse is shown in the histogram of Fig.~\ref{fig4}i. By summing up the contribution from all synapses we obtain a total energy consumption of 36mJ ("Memristor" row in Tab. \ref{tab1}). This value takes into account the four synaptic operations ($\Delta F$, Decay, W+F, and weight multiplications) of all memristive synapses of the entire STPN network for a whole Pong game. To give a nuanced comparison with a pure GPU implementation, we provide two separate measurements using an NVIDIA A100 40GB device (Method Section \ref{sec:GPU_energy} for details).
We report the median of 100 individual runs per synaptic operation for half- and single-precision floating-point arithmetic (fp16 and fp32, respectively).

First, we measure the GPU's energy consumption for executing each synaptic operation for all the network's $169984$ multi-functional synapses. The results are shown in the "GPU (standard)" row. It is observed that roughly one third of the total energy consumption stems from the three ST-Hebb specific operations ($\Delta F$, Decay, and W + F) and two thirds from the standard input-weight multiplication. 
We note that since the GPU is a massively parallel machine, this number of synapses may not fully utilize the device, potentially leading to lower energy efficiency. Indeed, the A100 GPU achieves the highest energy efficiency for a hypothetical network with around $2^{21}$ synapses. The case labeled "GPU (optimal)" is the corresponding energy consumption scaled to the original network's number of synapses. By comparing the fp16 case of the "GPU (optimal)" energy consumption with the total in the "Memristor" row an improvement of a factor of 96 is obtained. The saved power is due to both the multi-functional nature of our memristors and their in-memory compute capabilities, which in combination allow for the simultaneous computation of four operations without any memory traffic. The absence of memory traffic is especially beneficial, because all operations considered (i.e., element-wise and vector-matrix multiplication) have little to no data reuse and are memory-bound. As a consequence, most energy is consumed in data movement rather than computation (von Neumann bottleneck). This is demonstrated in Method Section \ref{sec:GPU_energy} where we quantify the energy consumption of the GPU's memory traffic: It accounts for more than 98\% of the total. Note that we did not include the power consumption of the required peripheral circuitry into our estimations for the memristor case. Nevertheless, the significant gain in energy efficiency and the simple two-terminal structure of our memristors are key indicators that a scaled up version of our memristive hardware is feasible. \\


\begin{table}[h!]
\caption{Energy consumed in mJ by the whole STPN network during one game of Atari Pong. The energy consumption is given for the operations that are unique to the ST-Hebb synapses ($\Delta F$, Decay, and W+F) as well as for the input-weight multiplication (weight mult.), which is common to all synaptic models.}
\label{tab1}%
\begin{tabular}{@{}llrrrrrr@{}}
\toprule
  & & $\Delta F$ & Decay & W + F & \textbf{ST-Hebb} & weight mult. & \textbf{Total [mJ]}\\
\midrule
Memristor & & 0.4   & 35.6  & -\footnotemark[1] & \textbf{36.0} & -\footnotemark[2] & \textbf{36.0}\\
\midrule
\multirow{ 2}{*}{GPU (standard)} & fp16 & 3477.6 & 3453.8 & 4656.6 & \textbf{11588.0} & 23192.5 & \textbf{34780.5}\\
& fp32 & 3406.5 & 3540.4 & 5168.1 & \textbf{12115.0} & 22817.3 & \textbf{34932.3}\\
\midrule
\multirow{ 2}{*}{GPU (optimal)} & fp16 & 572.9 & 550.8 & 653.9 & \textbf{1777.6} & 1686.3 & \textbf{3463.9}\\
& fp32 & 996.8 & 815.4 & 2732.5 & \textbf{4544.7} & 1747.3 & \textbf{6292.0}\\
\botrule
\end{tabular}
\footnotetext[1]{The long- and short-term components both affect the conductance of the same device so that the addition W+F does not need to be explicitly performed.}
\footnotetext[2]{The input-weight multiplication is computed via $I=G \cdot U$, which consumes power during the read operation. This power consumption is, however, already accounted for by the Decay term, which requires the application of a constant bias $V_{bias}$ to control the $\Lambda$ parameter. Here, the worst-case scenario is assumed ($V_{bias} = 0.6$V applied to all synapses). The electrical current due to this voltage can be read out, giving the result of the input-weight multiplication. A crossbar array configuration is necessary to enable the full vector-matrix multiplication of all inputs with all synapses.}
\end{table}

\FloatBarrier

\section{Conclusion}\label{sec13}

We presented a two-terminal memristor based on STO that is able to store and compute both long- and short-term synaptic weight updates, effectively collocating memory and computation as well as long- and short-term dynamics. In particular, we demonstrated control over the short-term decay time constant without the need for an additional electrical contact or complex control signals, which implements a form of intrinsic meta-plasticity. All these features are essential for neuromorphic circuit implementations, e.g., STPN networks, which outperform traditional artificial neural networks in large-scale, complex machine learning tasks such as Atari Pong. We contributed here to the development of these networks with the introduction of m-STPN units, increasing the reliability during training and highlighting the importance of long decay time constants. Finally, in simulation, we compared our memristor implementation of an STPN network to a GPU one and obtained a significant increase in inference energy efficiency by a factor of at least 96.

To fully realize our simulation concept in hardware, further work is needed: Firstly, our STO memristors should be converted to vertical structures to better control the distance between its metallic electrodes and allow for the creation of crossbar arrays. Secondly, the long-term retention of our memristors should be improved, while still preserving their short-term plasticity. It has been suggested that an oxide layer between the Pt electrode and STO could increase the retention of low conductance states \cite{Siegel2021}. Moreover, since we observed a significant impact of the decay time constant on the training performance, different decay models should be investigated for both long- and short-term components in STPN networks. The advancement of such neural networks inspired by biology holds the potential to significantly increase the performance of AI applications across diverse dynamic environments. Furthermore, multi-functional memristive synapses with intrinsic dynamics could function as a key enabling technology for the energy efficient hardware implementation of next-generation neural networks.

\section{Methods}\label{sec11}
\subsection{Device Fabrication}\label{sec:device_fab}
Both electrode stacks (Cr-Pt and Ti-Pt) were patterned using e-beam lithography and deposited by e-beam evaporation (Supplementary Figs. S1a and S1b). After deposition the whole device was subsequently annealed at 300 degrees for 20 min in Ar atmosphere (Supplementary Fig. S1c). This step causes a thermal oxide to form at the Ti-STO interface, leaving behind oxygen vacancies \cite{Li2016NanoscaleTi/SrTiO3}. Annealing also likely leads to diffusion of chromium into STO, doping the STO in the process \cite{LaMattina2008DetectionCrystals}. The device stack was finally encapsulated within 30nm of SiN using plasma enhanced chemical vapour deposition (PECVD) to protect against oxidation (Supplementary Fig. S1d).

\subsection{Experimental Setup}\label{sec:setup}
The quasi static I-V characteristics were measured with a Keysight M9601A Source Measure Unit. Voltage pulses were generated with a Keysight 33500 Arbitrary Waveform Generator. The current was fed through a DHPCA-100 trans-impedance amplifier from Femto and read out with a Rohde\&Schwarz RTE 1104 oscilloscope. 

\subsection{Modified STPN Model}\label{sec:mSTPN}
The equations describing the forward pass through an STPN layer follow \cite{Rodriguez2022Short-TermForget}:
\begin{align}
\textbf{G}^{(t)} &=  \textbf{W} + \textbf{F}^{(t)} \label{eq1} \\
\textbf{h}^{(t)} &=  \tanh(\textbf{G}^{(t)} \textbf{x}^{(t)}) \label{eq2} \\
\textbf{F}^{(t+1)} &=  \underbrace{\boldsymbol{\Gamma} \odot (\textbf{x}^{(t)} \otimes \textbf{h}^{(t)})}_{\Delta F} + \underbrace{\boldsymbol{\Lambda} \odot \textbf{F}^{(t)} }_\text{Decay} \label{eq3}
\end{align}
where bold letters denote matrices, $\odot$ element-wise multiplications and $\otimes$ outer products. The STPN layer model is parameterized by the long-term weight $\textbf{W}$, the Hebbian association strength $\boldsymbol{\Gamma}$, and the short-term decay parameter $\boldsymbol{\Lambda}$. During training these 3 parameters are learned using back propagation through time (BPTT). While $\textbf{W}$ directly controls the synaptic strength the $\boldsymbol{\Lambda}$ and $\boldsymbol{\Gamma}$ parameters define how the synaptic weight responds to stimuli, effectively implementing a form of meta-plasticity or "learning to learn". The plastic update of the synapse is modeled by Eq. \ref{eq3}. Equations \ref{eq2} and \ref{eq3} are adapted slightly from the original work in \cite{Rodriguez2022Short-TermForget} to reflect the specific implementation here.
In addition to Eqs. \ref{eq1} to \ref{eq3} the original STPN model also includes a form of normalisation on both the synaptic input  $x \rightarrow x_{eff} = \frac{x}{||W+F||}$ and the plastic weight $F \rightarrow F_{eff} = \frac{F}{||W+F||}$ (Supplementary Fig. S7a). This speeds up stochastic gradient descent during training. The normalization of $F$ leads to a modification of Eq. \ref{eq3} where the decay parameter $\Lambda$ becomes $\Lambda_{eff} = \frac{\Lambda}{||W+F^{(t)}||}$. As a consequence, the decay time constant changes at every time step, because $F^{(t)}$ varies over time. Such variations can lead to instabilities during training and they cannot be straightforwardly implemented on our memristors. Another important feature is that the decay time constant $\Lambda$ cannot be constrained to a certain range because $\Lambda_{eff}$ depends on the values of $W$ and $F$, which are unknown at the start of training. However, clamping $\Lambda$ is important as training becomes highly unstable if synapses reach values $\Lambda_{eff} > 1$. The solution adopted to circumvent this issue in the original formulation of  \cite{Rodriguez2022Short-TermForget} consisted of starting with small values of $\Lambda$ at the beginning of the training to ensure that $\Lambda_{eff}$ does not exceed 1. This has the disadvantage that the network only slowly learns longer decay time constants. By removing the normalization of the plastic weight $F$ and only normalizing the input (Supplementary Fig. S7b) in our modified STPN unit we achieve a better performance during training and also make the implementation on memristors feasible.

\subsection{Network Training} \label{sec:training}
We closely follow the training protocol established in \cite{Rodriguez2022Short-TermForget}. Concretely we use RLLib \cite{pmlr-v80-liang18b} to train and evaluate agents in PongNoFrameskip-v4. Preprocessing (dimensionality and color scale) for the game frames is done as in \cite{pmlr-v48-mniha16} with the exception of frame stacking, which was omitted. The training parameters were also adopted from \cite{Rodriguez2022Short-TermForget}: rollout length (50), gradient clipping (40), discount factor (0.99) and a learning rate starting at 0.0001 with a linear decay schedule finishing at $10^{-11}$ at 200 million iterations. Models are trained from the experience collected by 4 parallel agents.

\subsection{GPU Energy measurement} \label{sec:GPU_energy}
To fairly compare the efficiency of a memristor and GPU implementation of the network in Fig. \ref{fig4}a, it is essential that the GPU's energy consumption is only measured for the specific arithmetic operations that can be performed on the memristor: (1) W+F, (2) Decay, (3) $\Delta F$ and (4) weight multiplication (for more details see Supplementary Section S10).
We achieve this by running each operation separately with a specifically designed Python script. 
To measure the GPU's energy consumption, we use the pyJoules library \cite{Belgaid_Pyjoules_Python_library_2019}, which is a Python wrapper for NVIDIA's own energy reporting framework, NVIDIA Management Library (nvml). Since all operations have very short runtimes, to improve accuracy, we measure the energy spent for 10,000 to 200,000 executions of the corresponding kernels.
We report the median of 100 multi-executions and estimate the 99\% confidence interval (CI) using bootstrapping with 1000 samples. 
We report the GPU energy consumption of operations (1) to (4) in three ways: (I) per full Atari Pong game of the whole neural network, which employs $64 * 2656 = 169984$ synapses and runs for $6826$ time steps (Tab. \ref{tab1} in the main text), (II) per Operation, i.e., a single synapse and one time step (Tab. \ref{tab2} in this section), and (III) for a single synapse over the course of a full Atari Pong game, i.e., $6826$ time steps (Fig. \ref{fig4}g).

(I) For the "GPU (standard)" results in Tab. \ref{tab1}, the matrices \textbf{W}, $\textbf{F}^{(t)}$, $\textbf{x}^{(t)}$, and $\boldsymbol{\Lambda}$ required by operations (1) to (4) have the same size as in the neural network simulation. For the "GPU (optimal)" results, we increase the size of the matrices using the formula $(2592+64)\cdot k$, where $k$ is a power of two and ranges from $64$ (original network size) to $4096$.
We report the energy spent for $k=1024$, which exhibits the highest energy efficiency, scaled down to the original network's size (see Supplementary Fig. S10b).

(II) The "GPU (standard)" and "GPU (optimal)" rows in Tab. \ref{tab2} were obtained by dividing the values of Tab. \ref{tab1} by the number of operations executed during the whole game ($64 * 2656 * 6826$). The energy is given per floating point operation (flop) in pJ. Note that the weight multiplication is computed by a fused multiply-add (FMA) operation, which counts as two flops (one for addition and one for multiplication).

In the "GPU (compute)" row of Tab. \ref{tab2} we implement a CUDA kernel that only operates on data stored in registers without reading/writing from/to the GPU global memory. These results therefore measure the energy spent for the computation only, without the contribution of memory traffic. Concretely, each kernel execution performs as many arithmetic operations (addition, multiplication or FMA) as needed for one complete Atari Pong game. To increase accuracy, each measurement combines 10,000 kernel executions. As before, we report the median of 100 multi-executions and estimate the 99\% CI. The energy consumption per flop for the weight multiplication correspond to approximately 5.9 and 9.5 pJ/flop for half- and single-precision. This is in the same ballpark as measurements provided by NVIDIA and independent testing of the GPU's floating point unit (FPU)~\cite{nvidia-slides,bhalachandra2022understanding}, which validates our measurements.
By comparing the "GPU (compute)" results with the "GPU (optimal)" we observe that the memory traffic accounts for more than 98\% of the GPU's total energy consumption . This result shows the remarkable energy efficiency of the GPU's FPU and the benefit of reducing memory traffic. Note, however, that this particular GPU implementation would not be useful in practice, because the results of the kernel's computations are not accessible via the memory and can therefore not be used by a program running on the GPU. For this reason the highest efficiency GPU benchmark that corresponds to a working implementation is the fp16 energy measurement in the "GPU (optimal)" row.  

(III) For the GPU energy consumption of a single synapse shown in Fig. \ref{fig4}g we made use of the energy measurements per operation in the fp32 case of the "GPU (standard)" row in Tab. \ref{tab2}. It should be noted that the energy values in Tab.  \ref{tab2} were computed by first measuring the energy consumed by all synapses of the network in parallel and then divided by the number of synapses. This ensures that the massive parallelism of GPU's is utilized, although we're only interested in the energy consumption of a single synapse. The energy contributions per operation were then cumulatively summed for all time steps to obtain the time-series GPU data in Fig. \ref{fig4}g.

We note that the kernels utilize the GPU's regular FP cores rather than the tensor cores because the operations (W+F, Decay, $\Delta F$ and weight multiplication) do not compute matrix-matrix products.\\

The specifications of our test system are:\\

\noindent Hardware:
\begin{itemize}
    \item GPU: NVIDIA A100 with 40GB memory
    \item CPU: 2x AMD EPYC 7742 @ 2.25 Ghz (2 x 64/128 Physical/Logical Cores)
    \item RAM: 512 GB
\end{itemize}
            
\noindent Software:
\begin{itemize}
    \item Rocky Linux release 8.4
    \item Python 3.11.5
    \item Pytorch 2.2.0 dev20230913
    \item CUDA 12.1.1
\end{itemize}

\begin{table}[h!]
\caption{Energy consumed per floating point operation (flop) in pJ . The energy consumption is given for the operations that are unique to the ST-Hebb synapses ($\Delta F$, Decay, and W+F) as well as for the input-weight multiplication (weight mult.), which is common to all synaptic models. Note that the weight multiplication is computed by a fused multiply-add (FMA) operation, which counts as two flops (one for addition and one for multiplication). The other operations ($\Delta F$, Decay, and W+F) count as one flop.}
\label{tab2}%
\begin{tabular}{@{}llrrrrrr@{}}
\toprule
  & & $\Delta F$ & Decay & W + F & \textbf{ST-Hebb} & weight mult. & \textbf{Total [pJ]}\\
\midrule
Memristor & & 0.3   & 30.7  & -\footnotemark[1] & \textbf{31.0} & -\footnotemark[2] & \textbf{31.0}\\
\midrule
\multirow{ 2}{*}{GPU (standard)} & fp16 & 2997.1 & 2976.6 & 4013.2 & \textbf{9987.0} & 9994.1 & \textbf{19981.1}\\
& fp32 & 2935.9 & 3051.3 & 4454.1 & \textbf{10441.2} & 9832.4 & \textbf{20273.6}\\
\midrule
\multirow{ 2}{*}{GPU (optimal)} & fp16 & 493.7 & 474.7 & 563.6 & \textbf{1532.0} & 726.7 & \textbf{2258.7}\\
& fp32 & 859.1 & 702.7 & 2355.0 & \textbf{3916.8} & 753.0 & \textbf{4669.8}\\
\midrule
\multirow{ 2}{*}{GPU (compute)} & fp16 & 12.2 & 12.2 & 13.3 & \textbf{37.6} & 5.9 & \textbf{43.5}\\
& fp32 & 18.1 & 18.1 & 23.6 & \textbf{59.8} & 9.5 & \textbf{69.3} \\
\botrule
\end{tabular}
\footnotetext[1]{The long- and short-term components both affect the conductance of the same device so that the addition W+F does not need to be explicitly performed.}
\footnotetext[2]{The input-weight multiplication is computed via $I=G \cdot U$, which consumes power during the read operation. This power consumption is, however, already accounted for by the Decay term, which requires the application of a constant bias $V_{bias}$ to control the $\Lambda$ parameter. Here, the worst-case scenario is assumed ($V_{bias} = 0.6$V) applied to all synapses). The electrical current due to this voltage can be read out, giving the result of the input-weight multiplication. A crossbar array configuration is necessary to enable the full vector-matrix multiplication of all inputs with all synapses.}
\end{table}

\FloatBarrier

\backmatter

\bmhead{Acknowledgments}
We would like to thank the Operations Team of the Binnig and Rohrer Nanotechnology Center, especially Antonis Olziersky, Roland Germann, Ute Drechsler, and Diana Davila for the generous sharing of their immense fabrication knowledge. We would also like to thank Johannes Hellwig for sharing his insights on the STO switching mechanism and Dhananjeya Kumaar for inspiring the title of this work. Funding from the Werner Siemens Foundation, the SNSF Strategic Japanese-Swiss Science and Technology Programme under project metacross (grant number 214068), and the SNSF ALMOND Sinergia project (grant number 198612) is acknowledged. Finally, this work used computational resources from the Swiss National Supercomputing Centre (CSCS) under project 1119.

\bmhead{Author contributions}
 CW developed the concept of the paper, fabricated and measured devices, implemented, tested and trained the neural network. CW wrote the paper with input from all the authors. AZ performed the measurement of the GPU energy consumption. TZ developed the characterization setup and helped with device measurements. KP assisted with fabrication. MM and MK provided feedback on the theoretical device operation principle. TM wrote part of the abstract and introduction and gave guidance on the original version of the STPN network. ML supervised the project and helped on the structuring of the paper. AE supervised the project and led the study with inputs on numerous topics including the fabrication/characterization of the devices and the writing of the paper. CW and AE conceived the project.

\bmhead{Competing interests}
The authors declare no competing interests.

\bmhead{Data availability}
Device measurement data is available from the corresponding author on reasonable request. 

\bmhead{Code availability}
The repository with the m-STPN network source code can be found here:
https://bitbucket.org/weilen-mc/stpn/src/submission/ \\

The GPU energy calculations are available here: https://github.com/Nano-TCAD/SpikeDecay


\bibliography{references}

\end{document}